\documentclass[runningheads]{llncs}
\usepackage{float}
\usepackage[american]{babel}
\usepackage{graphicx}
\usepackage{booktabs}
\usepackage{multirow}
\usepackage{epstopdf}
\usepackage{subfigure}
\usepackage{bm}
\usepackage{cite}
\begin{document}
\title{Detection-Rate-Emphasized Multi-objective Evolutionary Feature Selection for Network Intrusion Detection}
\titlerunning{Detection-Rate-Emphasized Multi-objective Feature Selection}
%
\author{Zi-Hang Cheng\inst{1,2} \and
Haopu Shang\inst{1,2} \and
Chao Qian\inst{1,2}}
\authorrunning{Cheng et al.}
%
\institute{National Key Laboratory for Novel Software Technology, Nanjing University, China \and 
School of Artificial Intelligence, Nanjing University, China
\email{\{chengzh,shanghp,qianc\}@lamda.nju.edu.cn}}
\maketitle              
\begin{abstract}
Network intrusion detection is one of the most important issues in the field of cyber security, and various machine learning techniques have been applied to build intrusion detection systems. However, since the number of features to describe the network connections is often large, where some features are redundant or noisy, feature selection is necessary in such scenarios, which can both improve the efficiency and accuracy. Recently, some researchers focus on using multi-objective evolutionary algorithms (MOEAs) to select features. But usually, they only consider the number of features and classification accuracy as the objectives, resulting in unsatisfactory performance on a critical metric, detection rate. This will lead to the missing of many real attacks and bring huge losses to the network system. In this paper, we propose DR-MOFS to model the feature selection problem in network intrusion detection as a three-objective optimization problem, where the number of features, accuracy and detection rate are optimized simultaneously, and use MOEAs to solve it. Experiments on two popular network intrusion detection datasets NSL-KDD and UNSW-NB15 show that in most cases the proposed method can outperform previous methods, i.e., lead to fewer features, higher accuracy and detection rate.

\keywords{Network intrusion detection  \and Feature selection \and Multi-objective
optimization \and Multi-objective evolutionary algorithm.}
\end{abstract}
\section{Introduction}
With the development of the Internet, network intrusion appears more and more frequently, threatening cyber security. For example, the core information infrastructures of many organizations and institutions are always suffering from serious cyber attacks. In October 2016, DynDNS that provides dynamic DNS services was attacked by a large-scale DDoS, causing a lot of access problems to the websites. The attack made a large number of important websites paralyzed, e.g., Twitter was even unable to be accessed for about 24 hours. In modern warfare, government websites and military command systems are also usually paralyzed by large-scale cyber attacks.

Extensive works have been conducted to detect network intrusion, and most of them focus on building intrusion detection systems (IDS), which can be generally divided into signature-based detection and anomaly-based detection~\cite{DBLP:journals/ml/IglesiasZ15}. Signature-based IDS maintains a database of known attack signatures by experts, and compares an incoming network packet against the stored attack signatures, where a match will trigger an alert. As a signature is required for each attack, this kind of method can hardly detect new attacks. By learning from past behavior patterns, anomaly-based IDS generates a model of normal behavior, which is used to decide whether a received network packet is normal or attack. As the behavior of attacks evolves continuously, anomaly-based IDS is more promising than the signature-based one, and thus has received most of research attention. 

In the past few years, researchers have been exploring the use of machine learning techniques (e.g., logistic regression~\cite{moustafa2016evaluation}, decision tree~\cite{moustafa2016evaluation,DBLP:conf/icissp/SharafaldinLG18} and naive Bayes classifier~\cite{moustafa2016evaluation,DBLP:conf/icissp/SharafaldinLG18}) to build anomaly-based IDS. However, the computer network data is usually composed of a large number of features about the connections~\cite{DBLP:journals/cn/KhammassiK20}, where some of them may be redundant or noisy. Thus, feature selection is often required when applying machine learning to anomaly-based IDS. By removing the irrelevant and redundant features, feature selection can reduce computational cost and even further improve the classification performance. There are mainly three types of approaches for feature selection, i.e., filter, wrapper, and embedded. Filter methods evaluate the relevance of each feature separately, and select the top-ranked ones. Wrapper methods directly use the performance of the final classifier based on a feature subset as an objective function, and employ some optimization techniques to search for a good feature subset. Compared with filter methods, wrapper methods require more computational overhead, but can guarantee a better classification performance. Embedded methods integrate feature selection into the classifier training process.

For feature selection in network intrusion detection, previous works mainly employed the wrapper methods, which formulated feature selection as an optimization problem. As the resulting optimization problem is usually NP-hard, evolutionary algorithms (EAs)~\cite{back1996evolutionary,DBLP:books/sp/ZhouYQ19} inspired by natural evolution are often applied. EAs are a type of general-purpose heuristic optimization algorithms, which simulate the natural evolution process with variational reproduction (e.g., mutation and crossover) and superior selection. Khammassi and Krichen~\cite{DBLP:journals/compsec/KhammassiK17} considered feature selection in network intrusion detection as a single-objective problem, where the objective function was a weighted sum of classification accuracy and the reciprocal of the number of features, and used Genetic Algorithm (GA) to search for an optimal feature subset. Experiments on two network intrusion detection datasets KDD'99~\cite{Dua:2019} and UNSW-NB15~\cite{DBLP:conf/milcis/MoustafaS15} showed that it could achieve better performance than the previous feature selection methods without EAs. As feature selection naturally involves multiple objectives, e.g., minimizing the feature subset size and maximizing the accuracy, Multi-Objective EAs (MOEAs) have been employed to solve the corresponding formulation. For example, Khammassi and Krichen~\cite{DBLP:journals/cn/KhammassiK20} proposed a wrapper feature selection approach for IDS based on NSGA-II. They considered the feature subset size and the classification accuracy as objectives to optimize and used NSGA-II to solve it. It could make a good trade-off between the feature subset size and the accuracy, and achieved impressive success on three popular network intrusion detection datasets, NSL-KDD~\cite{DBLP:conf/cisda/TavallaeeBLG09}, UNSW-NB15 and CIC-IDS2017~\cite{DBLP:conf/icissp/SharafaldinLG18}. Note that there are also a lot of works on evolutionary feature selection, which are not specially for network intrusion detection. Please refer to~\cite{DBLP:journals/tec/XueZBY16,DBLP:journals/swevo/NguyenXZ20}.

Though previous feature selection methods for network intrusion detection have achieved good results, they often consider the accuracy and the number of features only in the problem formulation, while ignoring the detection rate, which is, however, very critical to the performance of an IDS. The detection rate is the proportion of real attacks that are actually detected. In real-world applications, an IDS with high accuracy but low detection rate is meaningless, which will miss a large number of real attacks and bring immeasurable loss to the network system. Therefore, for network intrusion detection, it is more necessary to consider the detection rate rather than just focusing on the accuracy. 

In this paper, by considering the detection rate explicitly, we formulate feature selection for intrusion detection as a three-objective optimization problem: minimizing the number of features, maximizing the classification accuracy and maximizing the detection rate of attacks, which can be solved by any existing MOEA. To examine the effectiveness of our three-objective formulation, we employ three typical MOEAs (i.e., NSGA-II~\cite{deb2002fast}, MOEA/D~\cite{DBLP:journals/tec/ZhangL07} and NSGA-III~\cite{deb2013evolutionary}) in the experiments. In the feature selection and final classification phase, multiple wrapper learners are tested, including CART decision tree~\cite{DBLP:books/wa/BreimanFOS84}, logistic regression (LR)~\cite{bishop2006pattern} and random forest (RF)~\cite{breiman2001random}. The empirical results on two commonly used network intrusion detection datasets NSL-KDD and UNSW-NB15 show that the solution set obtained under the three-objective modeling has clear advantage (i.e.,  fewer features, higher classification accuracy and detection rate in most cases) over previous methods. Furthermore, the comparison between the three-objective and bi-objective formulations using the same MOEA (i.e., NSGA-II, MOEA/D or NSGA-III) validates the advantage of considering the detection rate.

The rest of the paper is organized as follows. Section 2 introduces some related works about feature selection in network intrusion detection. Section 3 introduces our proposed method based on three-objective problem formulation. Experiments are presented in Section 4. Section 5 concludes the work.

\section{Related Work}
\subsection{Feature Selection in IDS}
As the number of features to describe the network connections can be very large, which usually involves redundancy and noise, feature selection plays an important role in the field of IDS. Due to its good performance, researchers focus on the wrapper feature selection method, which formulates feature selection as a subset selection problem and employs search algorithms to solve it. For example, frugal search algorithms like exhaustive search, Sequential Forward Selection (SFS) and Sequential Backward Selection (SBS) can be directly used to search for a feature subset. SFS starts from an empty set, and greedily selects the best feature to add to the set in each round so that the accuracy of the wrapper classifier is maximized. Lee et al.~\cite{lee2017feature} performed feature selection using SFS and constructed an IDS with the selected feature subset using random forest. Experiments were conducted with the NSL-KDD dataset and the results indicated that feature selection was a necessary preprocessing step to improve the overall performance of IDS. Serpil et al.~\cite{ustebay2018intrusion} conducted Recursive Feature Elimination (RFE) by measuring the importance of features via information gain and deleting the least important feature iteratively. However, as the relationships among features are complex and the search space exponentially expands with the number of features, these methods can hardly obtain a satisfying feature subset. More powerful search algorithms are required.

EAs are a kind of random heuristic search algorithm inspired by natural evolution, which have shown great potential in subset selection~\cite{DBLP:conf/ijcai/0001Q0021,DBLP:conf/ijcai/0002Z022,DBLP:journals/ec/FriedrichN15,gu2023subset,DBLP:journals/tec/Qian20,DBLP:conf/aaai/0001BF20,DBLP:conf/ijcai/QianSYT17,DBLP:conf/nips/QianS0TZ17,DBLP:journals/ai/QianYTYZ19,DBLP:conf/nips/QianYZ15,DBLP:journals/ai/RoostapourNNF22,zhang2023sparsity}. Recently, EAs have been employed to solve the feature selection in IDS and achieved impressive success. For example, Vijayanand et al.~\cite{vijayanand2018intrusion} set the classification accuracy on the feature subset as the optimization objective and employed Genetic Algorithm (GA) to search for the optimal feature subset. However, the feature subset size was ignored in the optimization process, resulting in bad performance on the reduction of feature number. Khammassi~et~al.~\cite{DBLP:journals/compsec/KhammassiK17} set the weighted sum of classification accuracy and the reciprocal of the number of features as the optimization objective and used GA to solve it. However, how to balance the weights of classification accuracy and the number of features is not trivial, which has a great influence on the results.

\subsection{Search with MOEAs}
In recent years, some works formulated feature selection as a multi-objective optimization problem and solved it with MOEAs. Some of them focused on multi-classification and set multiple objective functions to represent the classification performance on each class separately. De~et~al.~\cite{de2014feature} proposed a multi-objective feature selection approach in IDS, using NSGA-II as the search algorithm and the Growing Hierarchical Self-Organising Maps (GHSOM) as the classifier. Specifically, it used the Jaccard coefficient of each class to measure the classification performance on multiple classes. Thus, the number of objectives was equal to the number of classes. Zhu~et~al.~\cite{zhu2017improved} was inspired by the work~\cite{de2014feature} and used an improved NSGA-III as a search strategy and GHSOM-pr as a classifier. They also used the Jaccard coefficient to measure the suitability of the selected feature subsets. However, since only the Jaccard coefficient of each class was considered in the fitness function, the feature subset size was ignored, which led to a limited overhead reduction. 

As feature selection naturally involves two objectives, i.e., minimizing the feature subset size and maximizing the accuracy, some works have employed MOEA to solve the corresponding two-objective formulation. For example, Khammassi and Krichen~\cite{DBLP:journals/cn/KhammassiK20} set the feature subset size and the classification accuracy as the optimization objectives and used NSGA-II to solve the feature selection problem in IDS. Compared with previous works, it has achieved impressive performance on the typical network intrusion detection datasets NSL-KDD, UNSW-NB15 and CIC-IDS2017. The summary of related work is shown in Table~\ref{tab1}.

Through experimental studies, we find that their method has poor performance on detection rate, which is another critical metric in the field of network intrusion detection, focusing on the frequency of missing attacks. In view of the above shortcomings, we propose to model the feature selection in network intrusion detection as a three-objective optimization problem, where feature subset size, classification accuracy and detection rate are optimized simultaneously, and use MOEAs to solve it.

\begin{table}[htbp]
  \centering
  \caption{Comparison of related work.}\label{tab1}
  \resizebox{\textwidth}{!}{
    \begin{tabular}{rlllr}
    \toprule
    \multicolumn{1}{l}{Work} & FS algorithm & Objective & Dataset & \multicolumn{1}{l}{Classifier} \\
    \midrule
    \multicolumn{1}{l}{ ~\cite{lee2017feature}} & SFS   & Accuracy & NSL-KDD & \multicolumn{1}{l}{RF} \\
    \multicolumn{1}{l}{~\cite{ustebay2018intrusion}} & RFE   & Accuracy & CICIDS2017 & \multicolumn{1}{l}{MLP} \\
    \multicolumn{1}{l}{~\cite{vasan2016dimensionality}} & PCA   & Accuracy & KDD CUP, UNB ISCX & \multicolumn{1}{l}{C4.5, RF} \\
    \multicolumn{1}{l}{~\cite{vijayanand2018intrusion}} & GA    & Accuracy & ADFA-LD, CICIDS2017 & \multicolumn{1}{l}{SVM} \\
    \multicolumn{1}{l}{~\cite{DBLP:journals/compsec/KhammassiK17}} & GA    & Accuracy, Number of features & KDD99, UNSW-NB15 & \multicolumn{1}{l}{C4.5, RF, NBTree } \\
    \multicolumn{1}{l}{~\cite{de2014feature}} & NSGA-II & Jaccard coefficient & DARPA, NSL-KDD & \multicolumn{1}{l}{GHSOM} \\
    \multicolumn{1}{l}{~\cite{zhu2017improved}} & NSGA-III & Jaccard coefficient & NSL-KDD, KDD'99 & \multicolumn{1}{l}{GHSOM-pr} \\
    \multicolumn{1}{l}{~\cite{DBLP:journals/cn/KhammassiK20}} & NSGA-II & Accuracy, & NSL-KDD, UNSW-NB15,  & \multicolumn{1}{l}{C4.5, RF, NBTree } \\
          &       & Number of features & CIC-IDS2017 &  \\
    \midrule
          & NSGA-II/ & Accuracy,  &       &  \\
    \multicolumn{1}{l}{Our method} & NSGA-III/ & Number of features,  & NSL-KDD, UNSW-NB15 & \multicolumn{1}{l}{CART, LR, RF} \\
          & MOEA/D & Detection rate &       &  \\
    \bottomrule
    \end{tabular}%
    }
\end{table}%

\section{The Proposed Three-objective Evolutionary Feature Selection Framework}
In this section, we formulate the feature selection in IDS as a three-objective optimization problem, where the feature subset size, classification accuracy, and detection rate are optimized simultaneously. Under this formulation, we use MOEAs to solve it, and propose the overall framework of our method DR-MOFS.

\subsection{Three-objective Formulation}
\begin{figure}[htbp]
\centering
\includegraphics[scale=0.3]{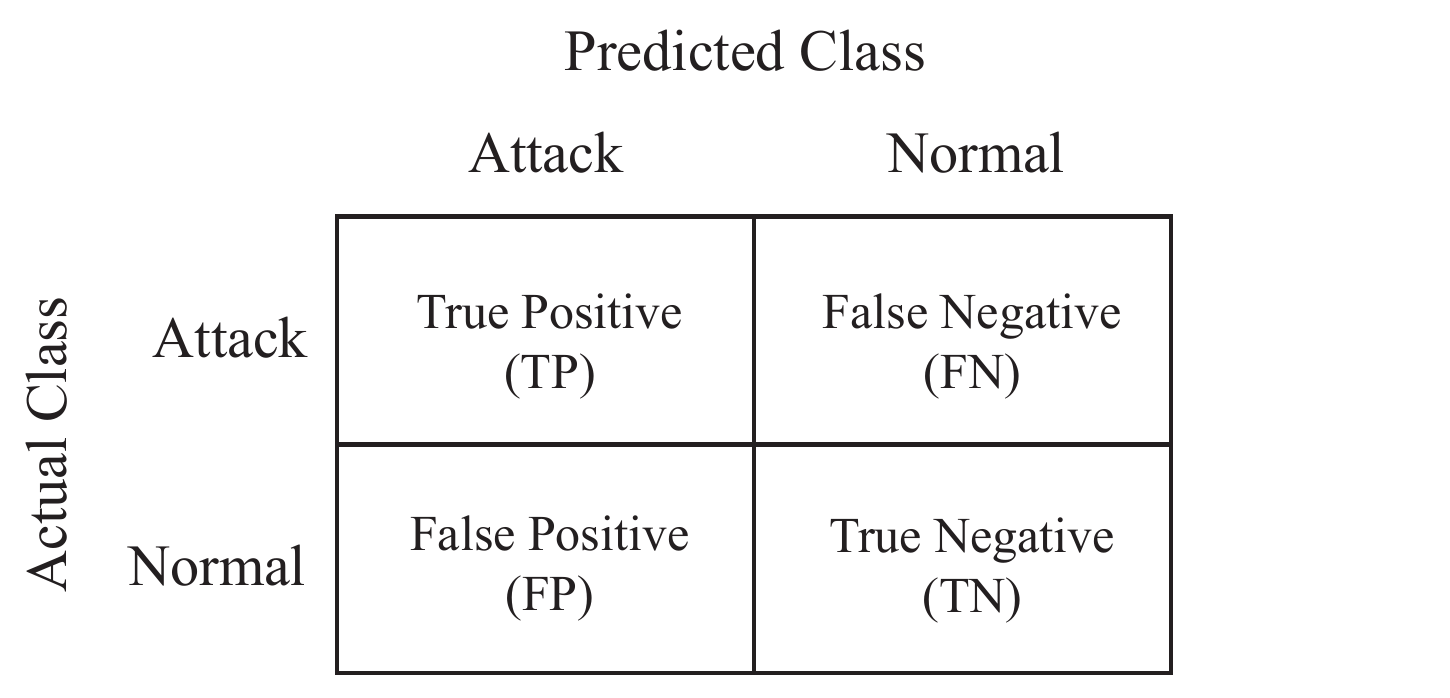}
\caption{Confusion matrix} \label{confusion matrix}
\end{figure}

Feature selection aims to search for a feature subset from all the original features, which can achieve better performance on the optimization objectives. Based on the observation that previous methods only consider the feature subset size and classification accuracy as the objectives to optimize, which have poor performance on detection rate, another critical metric in the special field of IDS, we naturally propose the three-objective optimization framework. That is, optimizing the three objectives simultaneously, i.e., minimizing the feature subset size, maximizing classification accuracy and maximizing detection rate, which can be generally formulated as
\begin{equation}
    \max_{\bm{x} \in {\{0,1\}}^n}\quad(\mathcal{-}\textrm{Size}(\bm{x}),\textrm{Accuracy}(\bm{x}),\textrm{DR}(\bm{x})),
\end{equation}
where a feature subset is represented by a binary string $\bm{x} \in {\{0,1\}}^n$, in which $x_i = 1$ means the corresponding feature is selected, otherwise it's not selected, and $n$ is the total number of features. $\textrm{Size}(\bm{x})$ denotes the number of features in the feature subset $\bm{x}$. Given a feature subset, we can evaluate its quality through the performance of a classifier trained on the feature subset. The classification result of a classifier trained on the feature subset $\bm{x}$ can be represented by the confusion matrix in Figure~\ref{confusion matrix}. The classification accuracy is denoted as $\textrm{Accuracy}(\bm{x})$, which describes the overall performance of a classifier. Under the notation in the confusion matrix, it is calculated as 
\begin{equation}
    \textrm{Accuracy}(\bm{x})=\frac{TP+TN}{TP+FN+FP+TN}.
\end{equation}
The detection rate of a classifier is denoted as \textrm{DR($\bm{x}$)}, which reflects the recall rate of attack samples. \textrm{DR($\bm{x}$)} is calculated as
\begin{equation}
    \textrm{DR}(\bm{x})=\frac{TP}{TP+FN}.
\end{equation}

Compared to the previous methods which mainly take feature size and classification accuracy as the objectives to optimize, we additionally incorporate detection rate as an optimization objective. This is because, in the field of network intrusion detection, rather than the normal traffic, we are more concerned about the anomalous traffic that may trigger network attacks, especially on how many of them are correctly identified and alerted by the IDS, since even one missing attack may cause severe loss. In addition, the data of network intrusion detection is usually imbalanced, which means that the amount of normal traffic data is much more than the anomalous traffic data, making the classification accuracy a less reliable metric to measure the detection ability of an IDS towards attacks. In other words, as the number of attacks is very small, an IDS that only identifies all the traffic as normal traffic can still have a very high accuracy, which is not appropriate. Therefore, the detection rate of anomalous traffic should be emphasized in feature selection. Since our formulation considers detection rate additionally, this allows MOEAs to discard solutions that perform well only in classification accuracy but poorly in detection rate during the search process. The existence of such solutions is the main reason why previous methods perform poorly in detection rate but still have a high classification accuracy.

\begin{figure}[htbp]
\includegraphics[width=\textwidth]{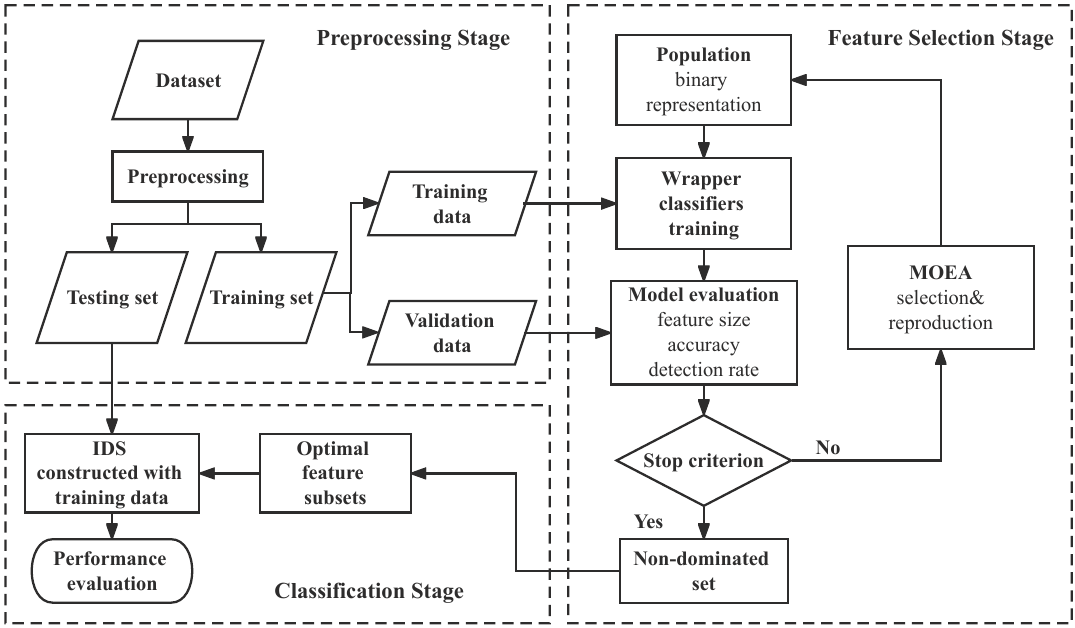}
\caption{The framework of the proposed detection-rate-emphasized multi-objective evolutionary feature selection method for IDS.} \label{framework}
\end{figure}
\vspace{-5mm}

\subsection{Overall Framework}
In this subsection, we introduce the overall framework of building an IDS with our detection-rate-emphasized multi-objective evolutionary feature selection. As shown in Figure~\ref{framework}, the whole process can be divided into three stages: preprocessing stage, feature selection stage and classification stage. In the preprocessing stage, firstly, we clean the original network traffic data to ensure that there are no missing values or infinity values in the dataset. After that, the categorical features are converted into numerical features by ordinal encoding. Finally, the features are normalized for the training of subsequent classifiers. The MOEAs are used in the feature selection stage to optimize the feature subsets which are represented using binary strings. Typical MOEAs such as NSGA-II~\cite{deb2002fast}, NSGA-III~\cite{deb2013evolutionary} and MOEA/D~\cite{DBLP:journals/tec/ZhangL07} can be directly used. In the classification stage, the optimized feature subsets are used to train the classifiers of IDS for the final evaluation and application.

Specifically, in the MOEAs, each individual represents a feature subset, which is encoded into a binary string. The individuals of the initial population are generated randomly, where each bit of an individual is set to 0 or 1 randomly with the same probability $0.5$. Given two individuals as parents, when generating the offspring individuals, we first employ the uniform crossover operator on the parents with the probability $p_\mathrm{c}$. Detailedly, the uniform crossover operator will generate two new individuals. Each bit in the binary string of the first individual is generated by randomly choosing a parent and copying the corresponding bit. The other individual is generated by copying the bits that are not chosen for the first individual. After that, the resulting two individuals are mutated with probability $p_\mathrm{m}$. The bit-wise mutation operator is used, where each bit independently has the same probability $1/n$ to flip and $n$ is the length of the binary string. When evaluating a new individual, a separate classifier is trained based on the corresponding feature subset and its classification accuracy and detection rate on the validation data are calculated, which are considered together with the feature subset size as the fitness of the individual. The selection strategy depends on the actually used MOEA. After the termination condition is reached, the algorithm will return a set of Pareto optimal feature subsets as the final solutions, which will be evaluated in the classification stage.

\section{Experiments}
In this section, we conduct experimental studies on the proposed detection-rate-emphasized feature selection method for IDS and compare it with previous feature selection methods. The experiments are conducted on two popular intrusion detection datasets, NSL-KDD~\cite{DBLP:conf/cisda/TavallaeeBLG09} and UNSW-NB15~\cite{DBLP:conf/milcis/MoustafaS15}. The datasets, contrast methods, and experimental comparisons are described in detail below.

\subsection{Dataset Description and Preprocessing}
Two typical public network intrusion datasets NSL-KDD and UNSW-NB15 are used to evaluate our method. NSL-KDD is the updated version of the classic network intrusion dataset NDD'99, which has wiped off some of the inherent problems and cleaned the invalid and redundant data. The dataset partition of the training set and testing set given by the official documentation is used in our experiments, which contains 125972 and 22543 instances respectively with four main attack types: DoS, Probe, R2L, and U2R. There are 41 features in each instance to describe the network connections. For convenience, we model it as a binary classification problem with normal and attack.

The UNSW-NB15 dataset was created by the cyber security research group at the Australian Centre for Cyber Security (ACCS)~\cite{DBLP:conf/milcis/MoustafaS15}. The created features can be classified into five categories: flow features, basic features, content features, time features, and additionally generated features. We also use the official data partition of the training set and testing set in our experiments, which contains 175341 and 82332 items respectively. Each instance contains 42 features. Similarly, we model the IDS as a binary classifier that identifies normal and attack classes.

During the data preprocessing stage, we remove instances with missing and infinite values in the datasets, encode categorical features into numerical features with an ordinal encoder, and finally normalize all attributes to be between 0 and 1. We randomly divide one-fifth of the training set as the validation set.

\subsection{Settings of Experiments}
Compared with the previous evolutionary feature selection methods, the main difference in this paper is the problem formulation, where feature subset size, classification accuracy and detection rate are optimized simultaneously. As for the selection of specific MOEA, we use the typical NSGA-II to compare our method with the previous methods for feature selection in IDS. Then we compare our detection-rate-emphasized feature selection method with typical bi-objective and single-objective evolutionary feature selection methods. The single-objective method~\cite{vijayanand2018intrusion} sets classification accuracy as fitness function and is optimized by GA. The bi-objective method~\cite{DBLP:journals/cn/KhammassiK20} employs NSGA-II to search for Pareto optimal feature subsets, where the feature subset size and classification accuracy are set as the objectives.

For the feature selection methods based on evolutionary algorithms, as different crossover and mutation operators will affect the search ability of the algorithms, for fair comparison, we uniformly use the same operators to generate the offspring individuals in the experiments. According to the parameters adopted by a large number of related works, we use the uniform crossover with probability $p_\mathrm{c} = 0.9$, together with the bit-wise mutation operator with probability $p_\mathrm{m} = 1$, which is introduced in Section 3.2. Furthermore, the population size is set as $P=100$ and the maximum number of generations is set as $G=500$, which is sufficient for convergence.

Some non-evolutionary feature selection methods are also examined for comparison, including Sequential Forward Selection (SFS)~\cite{lee2017feature} and Recursive Feature Elimination (RFE)~\cite{ustebay2018intrusion}. Note that Principal Component Analysis (PCA)~\cite{vasan2016dimensionality} is one of the prominent dimensionality reduction techniques which is widely used in the network traffic analysis. Therefore, we also use PCA to reduce the dimension of feature space and compare with it. Since they need to manually specify the number $k$ of remained features, we try $k=5, 10, 15, 20, 25$ separately in the experiments and report the result with maximum accuracy.

\subsection{Experimental Results and Comparison}
The experimental results of our detection-rate-emphasized multi-objective feature selection method and other feature selection methods on the NSL-KDD dataset are shown in Table~\ref{tab2}. Since MOEAs return a non-dominated solution set where each solution represents a feature subset, we evaluate each feature subset in the testing phase. We use the solution with the highest accuracy on the testing set obtained by different methods for comparison. DR-MOFS represents the version of our method using NSGA-II, NSGA-II-2objs represents the previous two-objective optimization method that considers feature subset size and classification accuracy optimized by NSGA-II, and GA represents the method that only uses classification accuracy as the objective and optimized by GA. All the solutions are evaluated on three metrics: feature subset size (Size), accuracy and detection rate (DR). The method named basic represents the performance without feature selection. The methods using evolutionary algorithms are all repeated ten times, and the average results are shown. Three different classifiers are examined in the experiments, i.e., CART decision tree, logistic regression (LR) and random forest (RF), using default classifier parameters. It should be noted that under each experimental setting, the wrapper classifier in the feature selection stage and the classifier used in the classification stage are identical. 

\vspace{-3mm}
\begin{table}[htbp]
  \centering
  \caption{DR-MOFS vs. other methods on the solution with \textbf{max test accuracy} on dataset \textbf{NSL-KDD}. The best result in each column is bolded and ‘$\bullet/\circ$’ denotes DR-MOFS is significantly better/worse than the corresponding method by the t-test with confidence level 0.05. ‘w/t/l’ denotes the amount by which DR-MOFS is better, same or worse based on mean.}\label{tab2}
  \resizebox{\textwidth}{!}{
    \begin{tabular}{lcccccccccc}
    \toprule
    \multicolumn{1}{c}{\multirow{2}[4]{*}{Method}} & \multicolumn{3}{c}{CART Decision tree} & \multicolumn{3}{c}{Logistic Regression} & \multicolumn{3}{c}{Random Forest} & \multicolumn{1}{c}{\multirow{2}[4]{*}{\quad w/t/l \quad}}  \\
\cmidrule{2-10}          & Size  & Accuracy(\%) & DR(\%) & Size  & Accuracy(\%) & DR(\%) & Size  & Accuracy(\%) & DR(\%) \\
    \midrule
    DR-MOFS & \textbf{4.7±1.5} & \textbf{86.86±0.65} & \textbf{82.71±1.68} & \textbf{2.0±0.0} & \textbf{82.04±0.00} & \textbf{73.00±0.00} & \textbf{3.0±0.0} & \textbf{82.43±0.43} & \textbf{76.63±0.75} & - \\
    NSGA-II-2objs~\cite{DBLP:journals/cn/KhammassiK20} & 8.0±3.4$\bullet$ & 84.25±0.89$\bullet$ & 76.09±1.21$\bullet$ & 5.4±2.3$\bullet$ & 79.79±0.72$\bullet$ & 66.74±1.56$\bullet$ & 5.6±4.1 & 79.74±0.28$\bullet$ & 68.14±2.19$\bullet$ & 9/0/0\\
    GA~\cite{vijayanand2018intrusion} & 26.2±2.6$\bullet$ & 80.38±1.79$\bullet$ & 67.95±3.03$\bullet$ & 23.4±2.7$\bullet$ & 76.08±0.44$\bullet$ & 60.18±0.80$\bullet$ & 24.0±0.6$\bullet$ & 77.56±1.14$\bullet$ & 63.12±2.28$\bullet$ & 9/0/0 \\
    SFS~\cite{lee2017feature}   & 10$\bullet$    & 81.76$\bullet$  & 70.68$\bullet$  & 25$\bullet$    & 78.01$\bullet$  & 63.22$\bullet$  & 5$\bullet$     & 80.97$\bullet$  & 68.65$\bullet$ & 9/0/0 \\
    RFE~\cite{ustebay2018intrusion}   & 15$\bullet$    & 83.10$\bullet$  & 72.35$\bullet$  & 10$\bullet$    & 75.25$\bullet$  & 61.69$\bullet$  & 25$\bullet$    & 78.35$\bullet$  & 64.21$\bullet$ & 9/0/0 \\
    PCA~\cite{vasan2016dimensionality}   & 25$\bullet$    & 79.15$\bullet$  & 65.30$\bullet$  & 15$\bullet$    & 77.52$\bullet$  & 65.87$\bullet$  & 10$\bullet$    & 77.09$\bullet$  & 61.84$\bullet$ & 9/0/0 \\
    Basic(without FS) & 41$\bullet$    & 78.22$\bullet$  & 64.07$\bullet$  & 41$\bullet$    & 75.47$\bullet$  & 61.85$\bullet$  & 41$\bullet$    & 78.11$\bullet$  & 63.59$\bullet$ & 9/0/0 \\
    \bottomrule
    \end{tabular}%
    }
\end{table}%
\vspace{-3mm}

Experimental results on NSL-KDD show that our method can achieve a significant improvement in detection rate compared with the original features without selection and previous feature selection methods. Moreover, our method also outperforms the comparison methods on the other two metrics.

The experimental results on another dataset UNSW-NB15 are shown in Table~\ref{tab3}. It can be seen that the original IDS constructed on the original features without selection already has good performance on detection rate, but our method can still achieve the best results in most cases. In particular, our proposed DR-MOFS can always obtain better accuracy and detection rate than the two-objective method NSGA-II-2objs, which indicates the superiority of our three-objective formulation. So, our detection-rate-emphasized multi-objective evolutionary feature selection method can generally improve the performance of the IDS.

\begin{table}[htbp]
  \centering
  \caption{DR-MOFS vs. other methods on the solution with \textbf{max test accuracy} on dataset \textbf{UNSW-NB15}. The best result in each column is bolded and ‘$\bullet/\circ$’ denotes DR-MOFS is significantly better/worse than the corresponding method by the t-test with confidence level 0.05. ‘w/t/l’ denotes the amount by which DR-MOFS is better, same or worse based on mean.}\label{tab3}
  \resizebox{\textwidth}{!}{
    \begin{tabular}{lcccccccccc}
    \toprule
    \multicolumn{1}{c}{\multirow{2}[4]{*}{Method}} & \multicolumn{3}{c}{CART Decision tree} & \multicolumn{3}{c}{Logistic Regression} & \multicolumn{3}{c}{Random Forest} & \multicolumn{1}{c}{\multirow{2}[4]{*}{\quad w/t/l \quad}}  \\
\cmidrule{2-10}          & Size  & Accuracy(\%) & DR(\%) & Size  & Accuracy(\%) & DR(\%) & Size  & Accuracy(\%) & DR(\%) \\
    \midrule
    DR-MOFS & \textbf{8.8±3.8} & \textbf{87.09±0.32} & \textbf{96.31±0.57} & 13.7±1.9 & \textbf{80.99±0.03} & 99.43±0.16 & 11.5±3.7 & \textbf{87.65±0.23} & \textbf{97.66±0.19} & - \\
    NSGA-II-2objs~\cite{DBLP:journals/cn/KhammassiK20} & 10.3±2.4 & 86.95±0.29 & 96.14±0.38 & \textbf{11.2±0.4}$\circ$ & 80.93±0.03$\bullet$ & 99.40±0.03 & \textbf{9.3±2.6} & 87.36±0.20$\bullet$ & 97.03±0.21$\bullet$ & 7/0/2\\
    GA~\cite{vijayanand2018intrusion}    & 18.0±2.1$\bullet$ & 85.80±0.85$\bullet$ & 94.14±1.26$\bullet$ & 24.4±1.7$\bullet$ & 80.80±0.33 & 98.89±0.74$\bullet$ & 20.0±3.0$\bullet$ & 87.34±0.20$\bullet$ & 97.21±0.19$\bullet$ & 9/0/0\\
    SFS~\cite{lee2017feature}   & 15$\bullet$    & 86.91  & 95.70$\bullet$  & 25$\bullet$    & 80.84$\bullet$  & \textbf{99.83}$\circ$ & 25$\bullet$    & 87.63  & 97.62  & 8/0/1 \\
    RFE~\cite{ustebay2018intrusion}   & 10    & 86.16$\bullet$  & 95.38$\bullet$  & 15$\bullet$    & 80.98  & 99.34  & 25$\bullet$    & 87.62  & 97.65  & 9/0/0\\
    PCA~\cite{vasan2016dimensionality}   & 20$\bullet$    & 84.73$\bullet$  & 93.68$\bullet$  & 20$\bullet$    & 80.15$\bullet$  & 97.35$\bullet$  & 15$\bullet$    & 85.87$\bullet$  & 96.07$\bullet$  & 9/0/0\\
    Basic(without FS) & 42$\bullet$    & 86.18$\bullet$  & 95.88$\bullet$  & 42$\bullet$    & 80.05$\bullet$  & 97.25$\bullet$  & 42$\bullet$    & 87.55  & 97.54 & 9/0/0 \\
    \bottomrule
    \end{tabular}%
    }
\end{table}%

\begin{figure}[htbp]
\includegraphics[width=0.95\textwidth]{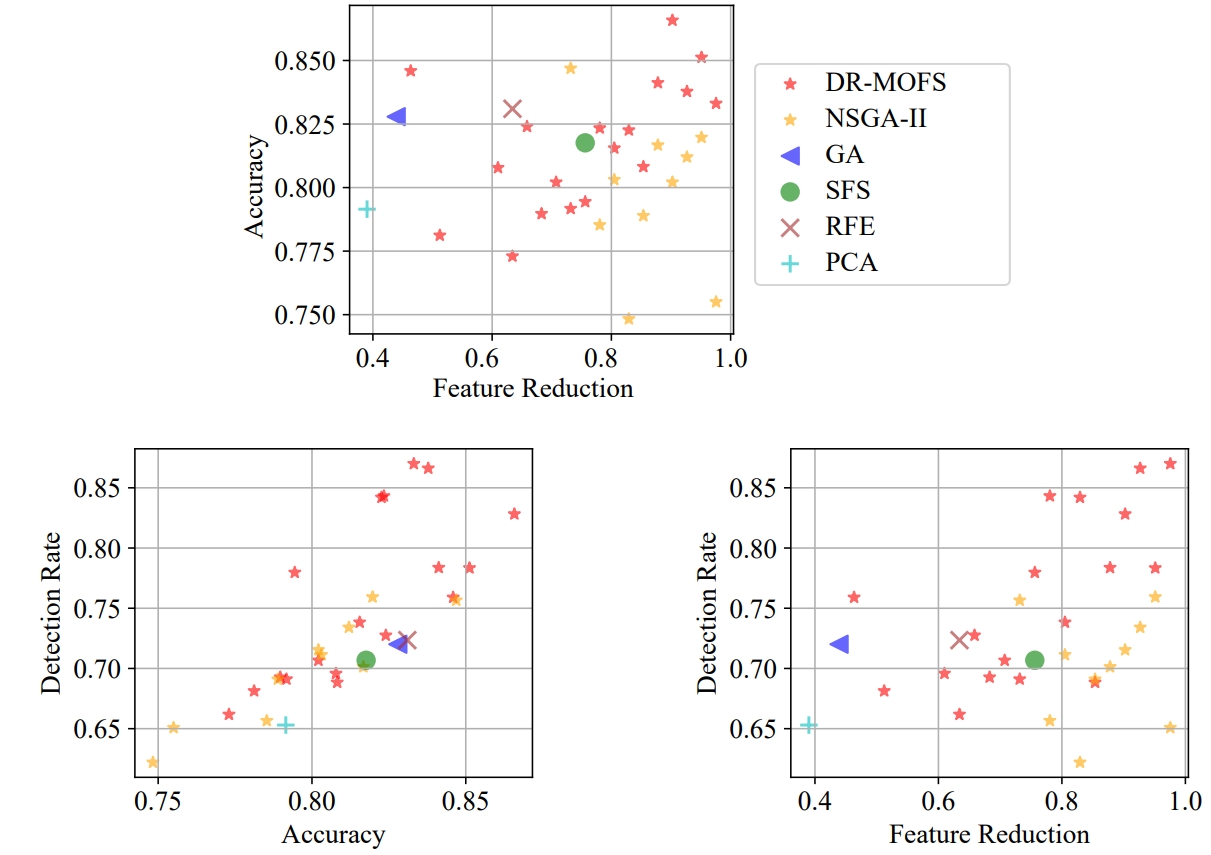}
\caption{The results of solutions obtained by different algorithms on three projection surfaces on NSL-KDD and CART.}
\label{projection-nsl}
\end{figure}

\begin{figure}[htbp]
\includegraphics[width=0.95\textwidth]{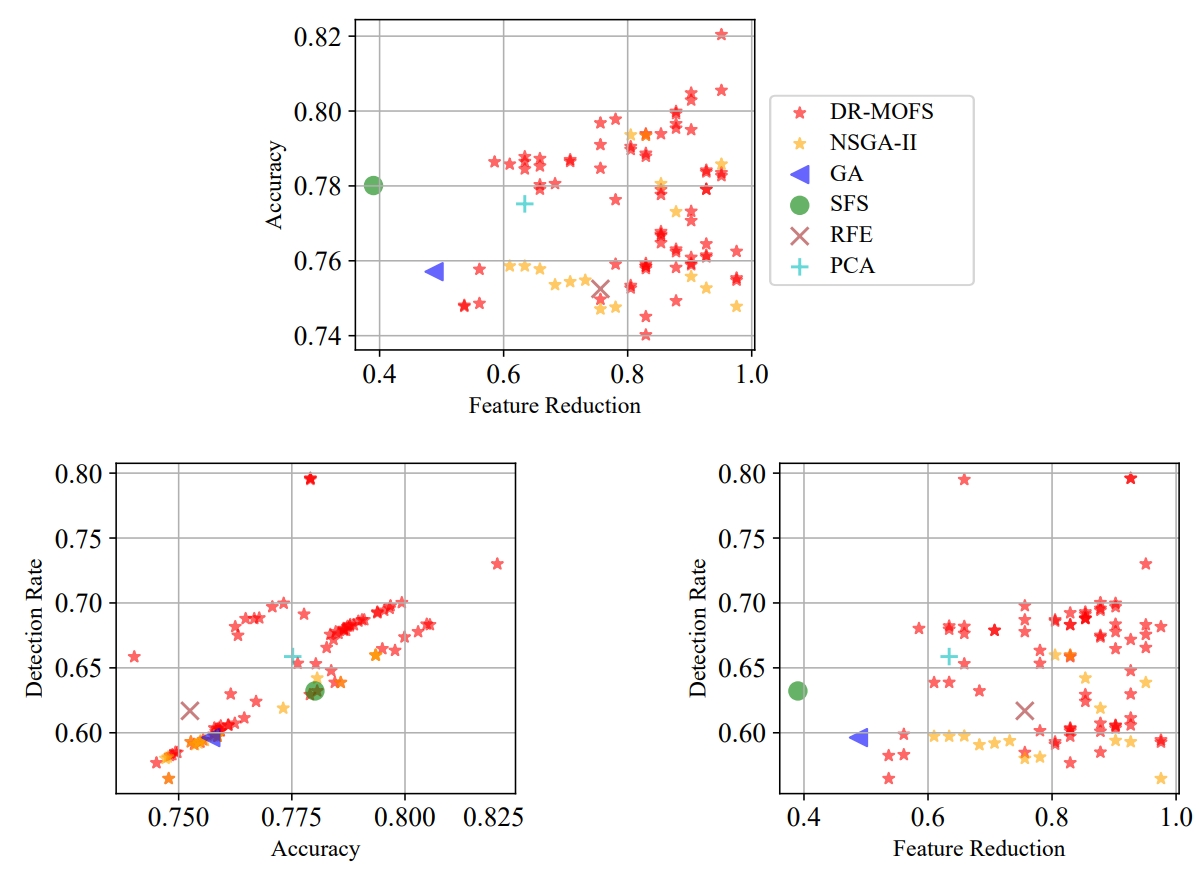}
\caption{The results of solutions obtained by different algorithms on three projection surfaces on NSL-KDD and LR.}
\label{projection-nsl-lr}
\end{figure}

The results in Table~\ref{tab2} and Table~\ref{tab3} only show the solution with the maximum test accuracy in each method, and may not be able to show the overall performance of the solution set returned by MOEAs. To precisely illustrate that our method can find solutions with better overall performance on each objective, we plot it as objective pairs in Figure~\ref{projection-nsl}. Here we replace feature subset size with feature reduction to maximize all the objectives (the closer to the upper right corner, the better), which is calculated as
\begin{equation}
    \textrm{Feature Reduction}(\bm{x})=1-\frac{\textrm{Size}(\bm{x})}{n}, 
\end{equation}
where $n$ is the total number of features. Because the Pareto dominance relationship between solutions in the 3D objective space is not intuitive to illustrate, we project the obtained solutions into the 2D objective space in pairs, as shown in Figure~\ref{projection-nsl}. This experiment is conducted using CART decision tree as the classifier on the NSL-KDD dataset. During the feature selection stage, DR-MOFS and NSGA-II-2objs obtain a set of solutions respectively, while other methods obtain one solution respectively. This explains why the methods using MOEAs have multiple points shown in Figure~\ref{projection-nsl}~(red stars and yellow stars). Note that we test the performance of all the solutions in the non-dominated solution set returned by NSGA-II, and then project them to the corresponding 2D surface. The red stars represent the results obtained by our method, some of which lie in the upper right corner, indicating that they have the best performance. Specifically, for each solution returned by the two-objective method NSGA-II-2objs and other comparison methods, there always exists at least one solution returned by our method that can dominate it. The results on the combination of NSL-KDD and LR are shown in Figure~\ref{projection-nsl-lr}.

All in all, a comprehensive comparison on the three objectives shows that our method consistently yields solutions that outperform the contrasting methods. In the real application, how to select the final feature subset to be used from a set of solutions should be determined according to the problem background and actual needs, and will not be further discussed here.

\subsection{Ablation Study}
In addition, to further demonstrate the advantages of our three-objective modeling over previous two-objective modeling of feature subset size and classification accuracy, we conduct ablation study on the objectives.

The first ablation study is based on the following research question: As one of the most common indicators for evaluating model performance in machine learning, F1-score comprehensively considers precision and recall and provides a balanced metric. So can we directly use the feature subset size and F1-score as two-objective optimization to achieve the same good results? Based on this, we replace the three-objective optimization problem (i.e., feature subset size, accuracy and detection rate) considered by DR-MOFS with a two-objective optimization problem (i.e., feature subset size and F1-score), and other components remain unchanged, represented by F1-MOFS. Table~\ref{tab4} and Table~\ref{tab5} are the results on the NSL-KDD and UNSW-NB15 datasets, respectively. As before, the solution with the highest test accuracy is shown for comparison. All solutions are evaluated based on the three metrics we care most about: feature subset size (Size), accuracy (Acc), and detection rate (DR). For each set of metrics, better results are marked in bold. The experiments are repeated ten times and the mean and standard deviation are recorded.

According to the ablation results, it can be observed that except for the slight advantage of F1-MOFS in the number of features obtained using logistic regression on the UNSW-NB15 dataset, DR-MOFS can always achieve better results under other experimental settings. In particular, for the dataset NSL-KDD in which the detection rate of the base learner is inherently low, DR-MOFS can relatively bring about a greater improvement in the detection rate. This demonstrates the advantages of DR-MOFS in optimization modeling.

\vspace{-3mm}
\begin{table}[htbp]
  \centering
  \caption{Comparison of DR-MOFS and its variant F1-MOFS on NSL-KDD.}\label{tab4}
  \resizebox{\textwidth}{!}{
    \begin{tabular}{lccccccccc}
    \toprule
    \multicolumn{1}{c}{\multirow{2}[4]{*}{Method}} & \multicolumn{3}{c}{CART Decision tree} & \multicolumn{3}{c}{Logistic Regression} & \multicolumn{3}{c}{Random Forest} \\
\cmidrule{2-10}          & Size  & Accuracy(\%) & DR(\%) & Size  & Accuracy(\%) & DR(\%) & Size  & Accuracy(\%) & DR(\%) \\
    \midrule
    DR-MOFS & \textbf{4.7±1.5} & \textbf{86.86±0.65} & \textbf{82.71±1.68} & \textbf{2.0±0.0} & \textbf{82.04±0.00} & \textbf{73.00±0.00} & \textbf{3.0±0.0} & \textbf{82.43±0.43} & \textbf{76.63±0.75} \\
    F1-MOFS & 10.6±5.1 & 82.45±0.21 & 73.41±1.53 & 4.4±2.0 & 79.92±0.81 & 66.96±1.79 & 3.2±1.0 & 79.58±0.06 & 68.33±2.52 \\
    Basic (without FS) & 41    & 78.22  & 64.07  & 41    & 75.47  & 61.85  & 41    & 78.11  & 63.59  \\
    \bottomrule
    \end{tabular}%
    }
\end{table}%

\vspace{-10mm}

\begin{table}[htbp]
  \centering
  \caption{Comparison of DR-MOFS and its variant F1-MOFS on UNSW-NB15.}\label{tab5}
  \resizebox{\textwidth}{!}{
    \begin{tabular}{lccccccccc}
    \toprule
    \multicolumn{1}{c}{\multirow{2}[4]{*}{Method}} & \multicolumn{3}{c}{CART Decision tree} & \multicolumn{3}{c}{Logistic Regression} & \multicolumn{3}{c}{Random Forest} \\
\cmidrule{2-10}          & Size  & Accuracy(\%) & DR(\%) & Size  & Accuracy(\%) & DR(\%) & Size  & Accuracy(\%) & DR(\%) \\
    \midrule
    DR-MOFS & \textbf{8.8±3.8} & \textbf{87.09±0.32} & \textbf{96.31±0.57} & 13.7±1.9 & \textbf{80.99±0.03} & \textbf{99.43±0.16} & \textbf{11.5±3.7} & \textbf{87.65±0.23} & \textbf{97.66±0.19} \\
    F1-MOFS & 12.4±0.8 & 86.79±0.21 & 95.81±0.29 & \textbf{11.4±0.8} & 80.91±0.00 & 99.38±0.00 & 15.2±2.3 & 87.56±0.10 & 97.36±0.16 \\
    Basic (without FS) & 42    & 86.18  & 95.88  & 42    & 80.05  & 97.25  & 42    & 87.55  & 97.54  \\
    \bottomrule
    \end{tabular}%
    }
\end{table}%
\vspace{-3mm}
Another ablation experiment is for the MOEA used for optimization. Besides NSGA-II, we further examine NSGA-III and MOEA/D, and then compare the three-objective and two-objective modeling using the same MOEA. The results on NSL-KDD and UNSW-NB15 are shown in Table~\ref{tab6} and Table~\ref{tab7}, respectively. The comparison is still based on the solution with the highest accuracy on the testing set, and three wrapper classifiers are examined. The experiments are divided into two groups, each containing three-objective and two-objective modeling with identical search strategy. For each group, the better results on different metrics are marked in bold. Experiments are repeated ten times and the mean together with standard deviation are shown in the tables. 

\vspace{-3mm}
\begin{table}[htbp]
  \centering
    \caption{Comparison of 3objs and 2objs using different MOEAs as search strategy on NSL-KDD.}\label{tab6}
  \resizebox{\textwidth}{!}{
    \begin{tabular}{cccccccccc}
    \toprule
    \multirow{2}[4]{*}{Methods} & \multicolumn{3}{c}{CART Decision tree} & \multicolumn{3}{c}{Logistic Regression} & \multicolumn{3}{c}{Random Forest} \\
\cmidrule{2-10}          & Size  & Accuracy(\%) & DR(\%) & Size  & Accuracy(\%) & DR(\%) & Size  & Accuracy(\%) & DR(\%) \\
    \midrule
    NSGA-III-3objs & \textbf{2.7±0.7} & \textbf{84.69±1.34} & \textbf{78.80±1.97} & \textbf{2.0±0.0} & \textbf{82.04±0.00} & \textbf{73.00±0.00} & \textbf{3.0±0.0} & \textbf{82.40±0.39} & \textbf{76.57±0.70} \\
    NSGA-III-2objs & 3.3±3.0 & 81.93±1.35 & 74.99±1.65 & 4.0±0.0 & 80.55±0.00 & 68.32±0.00 & 3.6±2.0 & 80.05±0.67 & 71.86±0.54 \\
    \midrule
    MOEA/D-3objs & \textbf{3.6±4.8} & \textbf{85.15±0.08} & \textbf{78.19±0.46} & \textbf{2.0±0.0} & \textbf{82.04±0.00} & \textbf{73.00±0.00} & 3.0±0.0 & \textbf{82.22±0.00} & \textbf{76.25±0.00} \\
    MOEA/D-2objs & 5.2±5.3 & 82.18±0.38 & 74.71±1.97 & 5.0±2.7 & 80.25±0.64 & 67.65±1.41 & \textbf{2.0±0.0} & 79.50±0.00 & 71.42±0.00 \\
    \bottomrule
    \end{tabular}%
    }
\end{table}%

\vspace{-10mm}

\begin{table}[htbp]
  \centering
  \caption{Comparison of 3objs and 2objs using different MOEAs as search strategy on UNSW-NB15.}\label{tab7}
  \resizebox{\textwidth}{!}{
    \begin{tabular}{cccccccccc}
    \toprule
    \multirow{2}[4]{*}{Methods} & \multicolumn{3}{c}{CART Decision tree} & \multicolumn{3}{c}{Logistic Regression} & \multicolumn{3}{c}{Random Forest} \\
\cmidrule{2-10}          & Size  & Accuracy(\%) & DR(\%) & Size  & Accuracy(\%) & DR(\%) & Size  & Accuracy(\%) & DR(\%) \\
    \midrule
    NSGA-III-3objs & \textbf{4.0±0.0} & \textbf{87.06±0.73} & \textbf{96.33±1.09} & \textbf{10.4±0.8} & \textbf{80.91±0.01} & \textbf{99.66±0.23} & \textbf{8.3±3.1} & \textbf{87.41±0.20} & \textbf{97.49±0.23} \\
    NSGA-III-2objs & 13.0±0.9 & 86.73±0.20 & 95.74±0.20 & 13.0±2.1 & \textbf{80.91±0.01} & 99.37±0.02 & 15.4±2.7 & 87.31±0.19 & 97.14±0.14 \\
    \midrule
    MOEA/D-3objs & 12.0±1.8 & \textbf{86.72±0.35} & \textbf{96.00±0.45} & 15.9±2.1 & \textbf{80.96±0.03} & 99.55±0.22 & 15.3±1.6 & \textbf{87.44±0.24} & \textbf{97.44±0.30} \\
    MOEA/D-2objs & \textbf{9.2±2.5} & 86.18±0.42 & 94.93±0.50 & \textbf{10.5±2.9} & 80.92±0.04 & \textbf{99.66±0.27} & \textbf{12.7±3.3} & 87.37±0.31 & 97.20±0.31 \\
    \bottomrule
    \end{tabular}%
    }
\end{table}%
\vspace{-3mm}
Obviously, except for the experimental combination of MOEA/D and LR on UNSW-NB15, the three-objective modeling can always have a better detection rate than the two-objective modeling. Meanwhile, in most cases the three-objective modeling can also obtain better accuracy, with a comparable reduction of feature subset size. The comparison of the ablation study under different MOEA settings illustrates the advantages of the modeling of DR-MOFS (i.e., feature subset size, accuracy, and detection rate) compared to the two-objective modeling (i.e., feature subset size and accuracy).

\section{Conclusion}
In this work, we propose DR-MOFS, a general multi-objective method for the important feature selection problem in the field of network intrusion detection, and employ MOEAs to solve it. Considering that previous methods only take feature subset size and classification accuracy as the objectives during the optimization process, leading to unsatisfying performance on a critical metric, i.e., detection rate, we analyze the necessity of emphasizing detection rate and propose the three-objective formulation, where feature subset size, classification accuracy and detection rate are optimized simultaneously. Experiments on two typical public network intrusion datasets NSL-KDD and UNSW-NB15, and three wrapper classifiers CART, LR, and RF show that our three-objective formulation can generally obtain better performance than the previous feature selection methods for IDS. The ablation study between two-objective modeling and three-objective modeling with different MOEAs further shows that our formulation can achieve a notably better detection rate with better classification accuracy and comparable feature subset size.

%
%
%
\bibliographystyle{splncs04}
\bibliography{ppsn.bib}


\end{document}